%% file: coling_latex.tex
\pdfoutput=1
\documentclass[11pt]{article}

\usepackage[]{coling}

\usepackage{times}
\usepackage{latexsym}

\usepackage[T1]{fontenc}
\usepackage[utf8]{inputenc}

\usepackage{microtype}

\usepackage{inconsolata}

\usepackage{multirow}
\usepackage{graphics}
\usepackage{array,graphicx}
\usepackage{float}
\usepackage{arydshln}
\usepackage{xcolor}
\usepackage{soul}
\usepackage{bm}

\usepackage{enumitem}
\usepackage{amssymb}% http://ctan.org/pkg/amssymb
\usepackage{pifont}% http://ctan.org/pkg/pifont
\newcommand{\cmark}{\ding{51}}%
\newcommand{\xmark}{\ding{55}}%

\title{Leveraging Large Language Models for Zero-shot Lay Summarisation in Biomedicine and Beyond}
\author{Tomas Goldsack$^{1}$, \textbf{Carolina Scarton}$^{1}$, \textbf{Chenghua Lin}$^{1,2}$ \\
        $^{1}$University of Sheffield, $^{2}$University of Manchester, \\
        \texttt{\{tgoldsack1, c.scarton\}@sheffield.ac.uk}\\
        \texttt{chenghua.lin@manchester.ac.uk}
}

\begin{document}
\maketitle

\input{sections/0.abstract}

\input{sections/1.introduction}
\input{sections/3.method}

\input{sections/4.experiments}

\input{sections/2.relatedwork}
\input{sections/6.conclusion}

\input{sections/7.limitations}

% Entries for the entire Anthology, followed by custom entries
\bibliography{custom}

\appendix
\input{sections/A0.appendix0}
\input{sections/A1.appendix1}
\input{sections/A2.appendix2}

\end{document}

%% file: sections/0.abstract.tex
\begin{abstract}

% Issues:

% Current Language Models (LLMs) have demonstrated remarkable performance in various generation-based tasks, even in a zero-shot setting. For Lay Summarisation, a task focusing on concisely explaining research articles to non-experts, their use has significant potential in making important research accessible to a broad audience. 

% Despite this potential, there is a lack of established knowledge regarding the performance of these models under different settings, raising questions about the optimal ways to utilise them. Additionally, the majority of previous Lay Summarisation studies have primarily concentrated on the Biomedical domain, leaving the generalisability of performance to other domains largely unexplored.
In this work, we explore the application of Large Language Models to zero-shot Lay Summarisation. We propose a novel two-stage framework for Lay Summarisation based on real-life processes, and find that summaries generated with this method are increasingly preferred by human judges for larger models. To help establish best practices for employing LLMs in zero-shot settings, we also assess the ability of LLMs as judges, finding that they are able to replicate the preferences of human judges. 
Finally, we take the initial steps towards Lay Summarisation for Natural Language Processing (NLP) articles, finding that LLMs are able to generalise to this new domain, and further highlighting the greater utility of summaries generated by our proposed approach via an in-depth human evaluation. 
\end{abstract}

%% file: sections/1.introduction.tex
\section{Introduction} \label{sec:introduction}

The goal of Lay Summarisation is to create a summary that effectively communicates the key concepts and findings of a technical article to a non-expert audience \citep{goldsack-etal-2022-making,zhang-etal-2024-atlas}. To achieve this, a well-written lay summary must, without hype or exaggeration, convey these findings clearly and concisely using everyday language, ensuring that it remains accessible to readers without specialised knowledge \citep{elifeDigest}. In recent years, the task of automatic Lay Summarisation has attracted increased interest in the research community, driven by the need to broaden access to scientific research. However, due to a lack of availability of publically accessible reference data, research for this task has been restricted to a few select domains, with Biomedicine being the most prominent.   

% In recent years, the task of automatic Lay Summarisation has seen increased interest, % cover significant recent works

\begin{figure}[t]
    \centering
    \resizebox{0.85\columnwidth}{!}{%
        \includegraphics{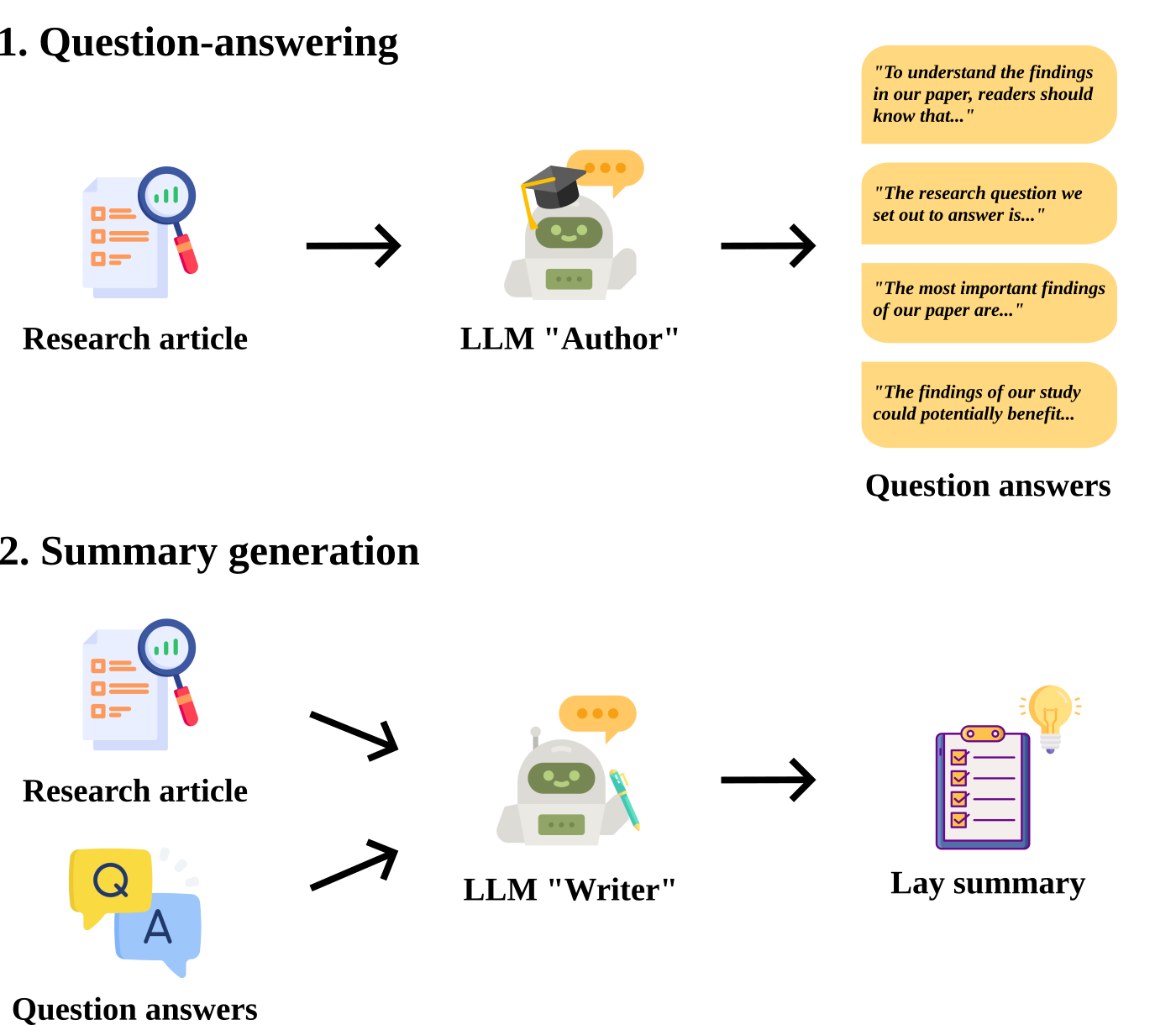}
    }
    \caption{A Visualisation of our two-stage Lay Summarisation framework, based on the real-life process of the eLife Journal.}
    \label{fig:framework}
\end{figure}

In recent years, the rise of Large Language Models (LLMs) has drastically altered the research landscape for Natural Language Generation (NLG), including for tasks such as Summarisation \citep{goyal2022,goldsack2023domain}. For Lay Summarisation specifically, the results of the recent BioLaySumm 2023 shared task suggest that LLMs have the potential to also push the boundaries of task performance \citep{goldsack-etal-2023-biolaysumm}, with two of the top three submissions adopting them within their approach \citep{turbitt-etal-2023-mdc, sim-etal-2023-csiro}.

At the forefront of this changing landscape are current LLMs' significantly improved zero-shot capabilities, emerging from the increased scale of models \citep{Kaplan2020ScalingLF} and breakthrough techniques such as instruction-tuning \citep{instruction-tuning}. Importantly, these zero-shot capabilities have significantly widened the scope of LLM applications, enabling us to explore tasks that were previously impossible due to a lack of training data. For Lay Summarisation, these models now provide the opportunity to explore domains outside of Biomedicine and improve accessibility to research in all domains. However, little is known about how best to utlise LLMs to generate lay summaries in a zero-shot setting, or how best to evaluate generated summaries in the absence of references. 

In this work, we attempt to address these questions and lay the groundwork for future Lay Summarisation research in unexplored domains. We propose a novel two-stage prompting framework for zero-shot Lay Summary generation inspired by real-world practices, before thoroughly evaluating the performance of our method in two domains: Biomedicine and Natural Langauge Processing (NLP). Our results show that summaries generated using the proposed framework are increasingly preferred by human judges as we increase the size of the underlying LLM, and that a panel of LLMs proves an effective proxy to human judges. Furthermore, we find that LLMs can effectively generalise to the previously-unexplored domain of NLP articles, and that the summaries generated with our proposed framework provide lay readers with a more well-rounded understanding of article contents in this domain.    

%% file: sections/3.method.tex
\section{Task Decomposition Framework} 

% Here we provide the details of our framework, emulating a realistic lay summary writing process whilst leveraging the key strengths exhibited by LLMs. 

% \subsection{Inspiration} \label{sec:inspo}
% % Explain how the original paper attests to the quality of these lay summaries
% % However, 

We explore the use of a two-stage framework for Lay Summarisation, based on the practices of the journal eLife.  
% a peer-reviewed journal dedicated to Biomedical and Life Sciences with an open-access approach, generates lay summaries (or "digests") for the research articles it publishes. 
In a recent study, \citet{goldsack-etal-2022-making} 
% utilised eLife articles to create a new dataset for Lay Summarisation. Through a comprehensive analysis, they found 
found that the lay summaries from eLife, crafted internally by journal editors and hired writers in collaboration with article authors, were notably more accessible than the author-written lay summaries from PLOS (the Public Library of Science).\footnote{eLife lay summaries are on average longer, more readable, and more abstractive than those of PLOS, positively impacting their comprehensibility.}
% Mention that we discussed this with their staff
%To ensure quality in the production of 
To produce their lay summaries, eLife adopts a multi-stage process involving different actors. \citet{elifeDigest} describe this process, whereby the paper authors are asked to answer a set of questions concerning the motivation and impact of their work, aiming to extract information that may not be explicit with the article itself but is essential to lay audience understanding.  
Upon inquiry with the team at eLife, we were provided with these questions, which inform our approach.

% \subsection{Implementation}

% As visualised in Figure \ref{fig:framework}, our framework decomposes the lay summary composition process into three simple stages. These stages cover the following: 1) an expert-based \textbf{question-answering} stage, whereby we extract important background information from the LLM using eLife-derived questions; 2) An initial \textbf{drafting} stage, where we utilise both the article and expert answers to construct a first draft of the output lay summary; and 3) a final \textbf{editing} stage, where the lay summary is assessed against a set of guidelines, and a final summary is produced. 
% In following this multi-stage approach and providing a simple task to each LLM actor, we aim to assess the extent to which LLMs can be used to emulate the process followed by eLife and what effect this has on the quality of generated summaries in comparison to standard one-stage generation. 
% More details of each of these stages are provided below:

As visualised in Figure \ref{fig:framework}, our framework is based on the introduction of an additional ``question-answering" stage in the summary generation process, with the answers generated from this stage used as an additional data source during summary generation. By introducing this additional stage, we aim to simulate the process that produces the high-quality lay summaries of eLife, and make explicit the important questions that should be addressed in the final summary. A detailed description of each stage is provided below:

\paragraph{1. Question-answering}
Given the article text, the LLM is asked to play the role of the paper author and answer four questions about the article. As previously mentioned, these questions are derived from those that eLife asks of authors 
% and are designed to extract information that a good quality lay summary should effectively communicate. 
Specifically, the questions are: 1) What background information would someone who is completely unfamiliar with your field need to know to understand the findings in your paper? 2) What exact research question did you set out to answer and why? 3) What are the most important findings of your paper? and 4) Who might eventually benefit from the findings of your study, and what would need to be done before we could achieve these benefits?\footnote{Alongside each question, additional guidance for answering is also presented. We provide a full explanation of this, and the prompts used for each agent, in Appendix \ref{sec:appendix1}.} 
% \footnote{Prompts for all agents are presented in Table \ref{tab:agent_response_prompts} in the Appendix.} 

 % Your summary should be between 300 and 400 words and containing minimal jargon, often using words and phrases that aren't present in the article. The first half of your summary should focus on explaining the background information that a lay audience will require, and the second half should explain the key experiments and results, finishing with a concluding sentence about the significance of the article.
 
\paragraph{2. Summary generation}
Given the article text and the ``author's" answers from the previous stage, we instruct the LLM to play the role of a writer and generate the lay summary based on guidelines derived from the analyses of \citet{goldsack-etal-2022-making}. These guidelines describe the desirable length, language, and structure of the generated summary.
% \paragraph{3. Editing}
% Finally, given the drafted lay summary and the same set of structural guidelines, the LLM is instructed to play the role of an editor and final revisions to the drafted summary. Importantly, the editor LLM is also provided with in-context input-output examples from the eLife train set to ensure the style, structure, and tone are replicated. Specifically, these examples consist of lay summaries generated using the first two stages of the framework (input) paired with the corresponding article references (output). % explain what this agent is useful for - e.g. self-correction + 

%% file: sections/4.experiments.tex
\begin{table*}[th]
    \centering
    \resizebox{0.85\textwidth}{!}{%
    \begin{tabular}{lcccccccccccccc} \hline
        \multirow{2}{*}{\textbf{Model}} &  \multirow{2}{*}{\textbf{\# Params}} & \multirow{2}{*}{\textbf{QA}} & \multicolumn{4}{c}{\textbf{Relevance}} && \multicolumn{2}{c}{\textbf{Readability}} && \multicolumn{1}{c}{\textbf{Factuality}} && \multicolumn{2}{c}{\textbf{H2H}}  \\  \cline{4-7} \cline{9-10} \cline{12-12} \cline{14-15}
        &&& \textbf{R-1$\uparrow$} & \textbf{R-2$\uparrow$} & \textbf{R-L$\uparrow$} & \textbf{BeS$\uparrow$} && \textbf{FKGL$\downarrow$} & \textbf{DCRS$\downarrow$} && \textbf{BaS$\uparrow$} && \textbf{PoLL$\uparrow$} & \textbf{PoH$\uparrow$} \\ 
                % \hline
                % BART && & 46.57 & 11.65 & 43.70 & 84.94 && 10.95 & 9.36 && -2.39   \\ 
                % Longformer &&& \textbf{47.23} & \textbf{13.20} & \textbf{44.44} & 85.11 && 11.88 & 9.09 && -2.56 \\  
                \hline
                \multirow{2}{*}{Phi2} & \multirow{2}{*}{2.7B} & \xmark & 41.66 & 9.10 & 39.39 & 83.27 && 13.08 & 9.05 && -3.06 && 0.65 & 0.80 \\
                 & & \cmark & 35.06 & 7.97 & 32.51 & 83.30 && 12.61 & 9.62 && -3.77 && 0.35 & 0.20  \\
                % \hdashline
                % Gemma-7B & 10.52 & 2.26 & 9.81 & 75.82 && 42.85 & 70.87 & \\
                % Gemma-7B$_{AW}$ & 26.45 & 4.63 & 25.03 & 78.30 && 9.20 & 7.71 &&  \\
                % \hdashline65057066169517
                \hdashline
                \multirow{2}{*}{Mistral} & \multirow{2}{*}{7B} & \xmark & 44.02 & 10.39 & 41.26 & 84.15 && 14.08 & 10.42 && -3.60 && 0.60 & 0.75 \\
                %  &  & \cmark & 39.61 & 9.19 & 37.11 & 82.39 && 15.86 & 9.47 && -3.54 && 0.40 & 0.25 \\
                % \hdashline
                &  & \cmark & 44.60 & 9.95 & 41.77 & 84.29 && 14.17 & 10.19 && -3.58 && 0.40 & 0.25 \\
                \hdashline
                
                \multirow{2}{*}{Mixtral} & \multirow{2}{*}{46B} & \xmark & 45.59 & 11.02 & 42.85 & 84.17 && 14.08 & 9.69 && -3.06 && 0.40 & 0.45 \\
                 && \cmark & 45.49 & 10.41 & 42.90 & 83.95 && 13.92 & 9.57 && -3.12 && 0.60 & 0.55\\
                \hdashline
                
                % \multirow{2}{*}{LLAMA2} &\multirow{2}{*}{70B} & \xmark & 42.77 & 10.79 & 39.93 & 84.41 &&  12.92 & 9.55 && -2.98 \\
                %  && \cmark  & 44.14 & 10.17 & 41.44 & 84.35 && 13.27 & 9.36 && -3.20 \\
                % \hdashline
                
                \multirow{2}{*}{LLAMA3} & \multirow{2}{*}{70B} &  \xmark  & 45.59 & 11.22 & 43.03 & 84.72 && 10.37 & 8.43 && -3.34 && 0.20 & 0.35 \\
                 && \cmark & 44.90 & 10.68 & 41.93 & 84.93 && 11.99 & 9.34 && -3.27 && 0.80 & 0.65 \\
                \hdashline
 
                % \multirow{2}{*}{QWEN} & \multirow{2}{*}{72B}& \xmark & \\
                % && \cmark  &  46.77 & 10.26 & 44.07 & 83.87 &&  12.12 & 9.77 &  & \\
                % \hdashline
 
                \multirow{2}{*}{DBRX} & \multirow{2}{*}{132B} &  \xmark & 44.77 & 11.09 & 42.04 & 84.25 && 13.48 & 9.69 && -3.10 && 0.15 & 0.30 \\
                 && \cmark & 44.18 & 9.94 & 41.54 & 84.29 && 14.31 & 10.17 && -3.24 && 0.85 & 0.70   \\
                \hline
    \end{tabular}}
    \caption{Average performance of models on eLife test split. \textbf{R} = ROUGE F1, \textbf{BeS} = BERTScore F1, \textbf{FKGL} = Flesch-Kincaid Grade Level, \textbf{DCRS} = Dale-Chall Readability Score, \textbf{BaS} = BARTScore, \textbf{PoLL} = Panel of LLm evaluators, \textbf{PoH} = Panel of Human evaluators.}
    \label{tab:results_elife}
\end{table*}

\section{Biomedical Lay Summarisation} \label{sec:experiment1}

To provide a comprehensive assessment of the performance of our proposed method on an existing dataset, 
we conduct several experiments using the test set of eLife dataset \citep{goldsack-etal-2022-making}, which contains 241 Biomedical research articles paired with expert-written lay summaries.
% Additionally, we compute the agreement of LLM judges in aligning with human preferences for lay summaries, ...
% \subsection{Experimental setup}

We compare our two-stage method against a standard one-stage lay summary generation prompt across several 
%LLMs before performing a multi-faceted exploration of prompt design, including an analysis of input selection and an ablation study. 
%We experiment with several 
popular open-source LLMs of various sizes: Phi2, Mistral-7B, Mixtral 8$\times$7B, LLAMA3-70B, and DBRX. 
% Note that both Mixtral and DBRX adopt the Mixture-of-Experts architecture. 
For each LLM, we generate lay summaries using both our proposed two-stage QA-based prompting method and a generic lay summary generation prompt: ``Generate a summary of the following article that is suitable for non-experts".\footnote{An additional discussion of prompts is given in Appendix \ref{sec:appendix1} and details of additional experiments, including an input selection analyses and ablation study, is given in Appendix  \ref{sec:appendix2}.} 

% We experiment with the use of two popular open-source LLMs, Mixtral \citep{jiang2024mixtral} and LLAMA2 \citep{touvron2023llama}.\footnote{Note that the same LLM is used for every stage within the framework.} Note that Mixtral and LLAMA models have 16384 and 4096 token input limits, respectively.  In addition to utlising these models within our proposed framework, we include three additional baseline variations: 1) LLM$_{zs}$ - a zero-shot variation; 2) LLM$_{fs}$ - a few shot variation; and 3) LLM$_{MDC}$ - a second-few shot variation based on the method proposed by \citet{turbitt-etal-2023-mdc}, the best-performing team of the recent BioLaySumm shared task \citep{goldsack-etal-2023-biolaysumm}. Using only the abstract as input, this method surrounds selecting the maximum number of in-context examples that can fit within the context window of the selected model and was tested using multiple GPT-family models (with \texttt{text-davinci-003} achieving the best performance).\footnote{The prompts used for all baselines are included in the Appendix.} Furthermore, to assess our proposed framework in both zero- and few-shot settings, we include two versions: 1) LLM$_{AW}$ -``Author-Writer" zero-shot variation (where the editing stage is excluded); and 2) LLM$_{AWE}$ - ``Author-Writer-Editor" few-shot variation.\footnote{All experiments were run locally using 4 A100 GPUs.}

\paragraph{Evaluation}
We adopt the automatic evaluation metrics used in the recent BioLaySumm Shared Task \citep{goldsack-etal-2023-biolaysumm}, assessing models along the dimensions of Relevance, Readability, and Factuality. Specifically, we include ROUGE-1, 2, and L \citep{lin-2004-rouge} and BERTScore \citep{bertscore} to measure relevance to references. For readability, we Flesch-Kincaid Grade-Level (FKGL) and Dale-Chall Readability Score (DCRS) are used. Finally, a version of  BARTScore \citep{bartscore}, adapted to process long inputs (using sparse attention) and fine-tuned on the eLife dataset, is used to measure Factuality. 

Finally, we perform a head-to-head comparison using a sample of 20 randomly selected test set instances for each LLM, using a panel of both lay human judges and LLM judges. Specifically, for a given sample instance, judges are provided with the lay summaries generated by each method and asked to decide which summary they find more useful as a layperson wanting to understand the findings and significance of the article.\footnote{To mitigate potential positional bias, we swap the order in which summaries are provided to judges randomly, such that 50\% of each ordering is used.} For LLMs judges, we follow \citet{verga2024replacing} and employ a panel of 3 smaller LLMs - namely Command R, Haiku, and GPT-3.5. For human judges, we employ 4 lay people with no experience in biomedicine.  
In both cases, judges indicates their preferences indiviually and preferences are then aggregated using majority vote.\footnote{When a ties in voting occurs betwen human judges, we take the preference of the judge who obtained the highest agreement with other judges over all instances. A further discussion on evaluator agreement is provided in Appendix \ref{sec:appendix2}.}
We report the proportion of each summary type preferred by both sets of judges.

\paragraph{Results and discussion}
% Here, we will provide the details and discussion for each experiment in turn. Specifically, in addition to comparing the best-performing configurations for each system, we conduct experiments to measure the effects of in-context learning, input selection, and prompt design on performance. 

% \paragraph{Overall system comparison}
% Things to discuss:
% 1. Comparison between Supervised, few-shot, and zero-shot
% 2. Comparsion between LLM baselines + our method (in both few shot and zero shot settings)
% 3. Comparions between the two different LLMs used (in all settings)

Table \ref{tab:results_elife} presents the metrics scores obtained by both the standard one-stage prompt and the proposed two-stage prompt for each LLM (where a \cmark the \textbf{QA} column denotes the two-stage prompt. Interestingly, we observe different patterns occuring for smaller ($\le$ 7B) and larger ($\ge$ 46B) LLMs. 
For the smaller LLMs - Phi2 and Mistral - we see a significant drop in the scores of automatic metrics when the proposed two-stage framework is used, particularly ROUGE scores. Additionally, we find that both human and LLM judges prefer the standard one-stage prompt in this setting. However, for larger LLMs - Mixtral, LLAMA3, and DBRX - we that, despite roughly comparable metric scores, judges tend to prefer summaries generated by the proposed two-stage framework. Furthermore, the extent to which judges prefer these summaries increases as model size increases.

Overall, these results indicate that: 1) the proposed two-stage framework produces lay summaries that are increasingly preferred by lay people as model size increases, but presents too complex a task for smaller LLMs; 2) automatic metrics such as ROUGE fail to capture some aspects of the summary that inform human preferences for higher quality summaries; and 3) the panel of LLM judges (PoLL) is proves quite effective in approximating human preferences in all cases.

\begin{table}[t]
    \centering
    \resizebox{0.9\columnwidth}{!}{%
    \begin{tabular}{lccccccc} \hline
        \multirow{2}{*}{\textbf{Model}} &  \multirow{2}{*}{\textbf{QA}} & \multicolumn{2}{c}{\textbf{Readability}} && \multicolumn{1}{c}{\textbf{Factuality}} && \multicolumn{1}{c}{\textbf{H2H}} \\  \cline{3-4} \cline{6-6} \cline{8-8} 
        & & \textbf{FKGL$\downarrow$} & \textbf{DCRS$\downarrow$} && \textbf{BaS$\uparrow$} && \textbf{PoLL$\uparrow$} \\ 
                \hline
                 \multirow{2}{*}{Mixtral} & \xmark & 13.88 & 9.44 && -3.17 && 0.40 \\
                & \cmark & 14.59 & 9.48 && -3.28 && 0.60 \\
                \hdashline
                \multirow{2}{*}{LLAMA3} & \xmark & 10.60 & 8.37 && -3.53 && 0.30 \\                
                & \cmark & 12.47 & 9.47 && -3.59 && 0.70 \\
                \hdashline
                \multirow{2}{*}{DBRX} & \xmark & 13.23 & 9.84 && -3.22 && 0.25  \\
                & \cmark & 14.91 & 10.07 && -3.34 && 0.75 \\
                \hline
    \end{tabular}}
    \caption{Average performance of models on ACL paper set. \textbf{FKGL} = Flech-Kincaid Grade Level, \textbf{DCRS} = Dale-Chall Readability Score,  
    \textbf{BaS} = BARTScore.
    }
    \label{tab:results_acl}
\end{table}

\section{Lay Summarisation for NLP} \label{sec:experiments}

% Due to their demonstrated zero-shot capabilities, LLMs can also be deployed for domains where we lack training data. Previous approaches to Lay Summarisation have almost exclusively applied to Biomedical research articles, with this being the only domain for which there exists gold-standard reference data. 
In this section, we explore the application of our framework to NLP, a domain with a high publication rate 
% and, due to recent breakthroughs resulting in viable consumer products like ChatGPT, has 
that has recently been gaining significantly more attention from non-expert audiences (i.e., the general public). 
% In the absence of reference lay summaries, we conduct a reference-less automatic evaluation, followed by a carefully designed human evaluation that utilises both lay people and experts to provide a comprehensive assessment of lay summary quality. 
% \paragraph{Experimental setup} 
We collect 100 randomly sampled articles from the proceedings of ACL 2023 to use as a test set. To delve deeper into the quality of summaries produced by each zero-shot prompt, we conduct a reference-less automatic evaluation, followed by a carefully designed human evaluation that utilises both lay people and experts to provide a comprehensive assessment of lay summary quality. 
% zero-shot models in both a one-stage and multi-stage setting,
% we assess the performance of Mistral-7b${AW}$ and Mistral-7b${zs}$ on this test set. Notably, both models exhibit comparable performance in summarizing eLife content.  

% use the best-performing model
% as a baseline, use the same model in a few-shot setting (or maybe zero shot, or both)

\paragraph{Evaluation}  
In the absence of reference lay summaries, we repeat our evaluation protocol from Table \ref{sec:experiment1} using only reference-less metrics. 
% we can get an idea of the consistency of our models across domains.
% \paragraph{Human Evaluation} 
To gain further insight into the specific differences between summaries generated in each setting, we use our lay human evaluators to perform an alternative human evaluation using a random sample of 20 articles from the ACL paper set, carefully designed to assess how the summaries differ in their utility.
Specifically, our 4 lay participants are tasked with answering the following set of questions about the article, based on only the lay summary: 1) What problem is the article tackling? 2) How did the authors tackle the problem? 3) What are the key findings of the article? 4) Why are these findings significant?\footnote{To avoid any cross-contamination of knowledge between generated summaries, each lay annotator only assesses 1 generated summary per article. To ensure a fair comparison between models, our 4 annotators are split into 2 groups of 2, with each group assessing 10 summaries generated by each model.} Subsequently, we employ 3 NLP experts to judge the answer's of lay participants by classify an answer based on the extent to which they agree with it.\footnote{All expert judges are PhD students studying NLP.} This is done using the labels ``Completely agree", ``Somewhat agree", ``Somewhat disagree", and ``Completely disagree". 

% Our evaluation can be split into two stages:
% 1) We recruit 4 participants with no technical experience (i.e., lay people) to answer a set of questions about the article based only on the generated lay summary, and 2) we recruit 3 NLP experts to judge the answers. Specifically, expert judges have to classify an answer based on the extent to which they agree with it using the labels ``Completely agree", ``Somewhat agree", ``Somewhat disagree", and ``Completely disagree". 
% Specifically, we ask evaluators to answer the following four questions:  
% In combination, these questions encapsulate the essential knowledge a layperson should be able to derive from a lay summary.

% Experts have to mark answers with:
% - Completely agree
% - Somewhat agree
% - Somewhat disagree
% - Completely disagree

% Perhaps one way to evaluate these would be to have humans (lay people) rank summaries produced by different models/methods

% Another way would be to have them answer questions based on the lay summary content

% \begin{figure}
%     \centering
%     \resizebox{0.9\columnwidth}{!}{%
%     \includegraphics{emnlp2023-latex/figures/human_eval_results.png}
%     }
%     \caption{Human evaluation results.}
%     \label{fig:human_eval_results}
% \end{figure}

\begin{figure}
    \centering
    \resizebox{0.85\columnwidth}{!}{%
    \includegraphics{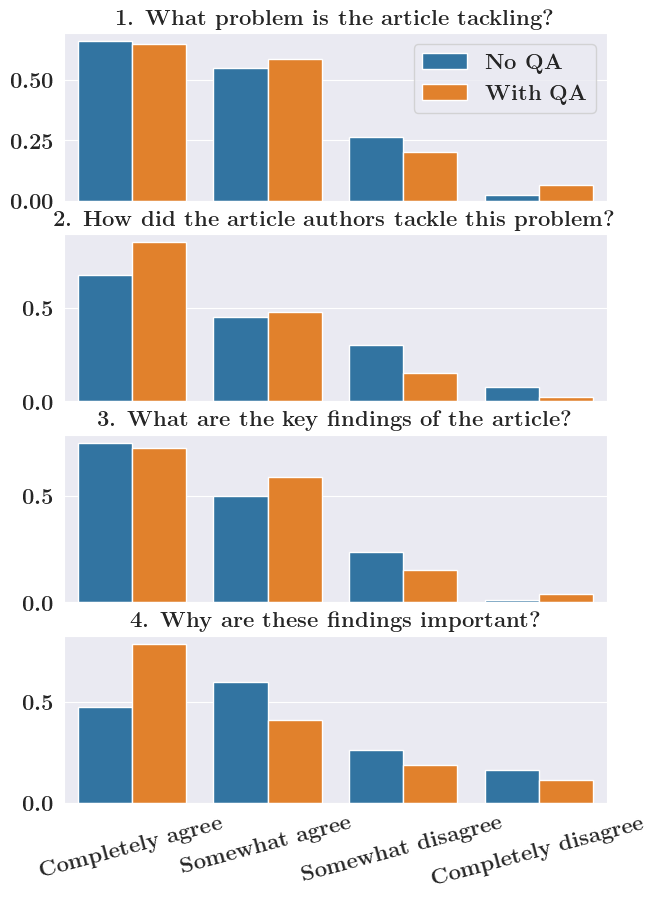}
    }
    \caption{Human evaluation results as the proportion of answer votes for each question.}
    \label{fig:human_eval_results}
\end{figure}

\paragraph{Results and Discussion}
Table \ref{tab:results_acl} presents the automatic metric scores obtained by the larger LLMs, for which human judges preferred summaries generated by the proposed framework. We find that scores obtained for NLP articles appear largely comparable with those obtained for biomedical articles (Table \ref{tab:results_elife}), suggesting that both approaches generalise to the NLP domain.

Figure \ref{fig:human_eval_results} presents the results of our human evaluation, visualising the total proportion of votes both models received for each classification label.
Firstly, it is evident that the answers of lay participants for both methods receive overwhelmingly more positive classifications (agree) than negative classifications (disagree) from expert annotators, suggesting that both methods produce reasonably good lay summaries when applied to NLP articles.
However, we find that the two-stage approach (with QA) almost consistently receives fewer votes for the ``disagree" categories across all questions (except for question 1), indicating that these summaries are generally better in enabling laypeople to provide answers that demonstrate some level of understanding.
Furthermore, particularly significant gains for positive labels can be seen for questions relating to article methodology and significance (2 and 4, respectively; suggesting that answers extracting during the QA stage of the framework are particularly pertinent for these aspects.

%% file: sections/6.conclusion.tex
\section{Conclusion} \label{sec:conclusion}

In this work, we study the utility of LLMs for zero-shot Lay Summarisation and proposing a novel two-stage framework inspired by professional practices.
Our experiments on biomedical articles suggest that the benefit provided by our framework increases in line with the size of the LLM, as demonstrated by increased preference of lay human evaluators. Although standard automatic metrics such a ROUGE fail to capture such preferences, we also find that a Panel of LLM evaluators proves effective. Our experiments on NLP article attest to the ability of models to generalise to an unexplored domain, and further illustrate how the proposed approach enables a more well-rounded understanding of article contents.
% of article contents and highlighting limitations in automatic metrics in capturing such aspects.  

% relating to how input selection, in-context learning, and prompt design affect performance. Finally, we conduct the first study on the

%% file: sections/7.limitations.tex
\section{Limitations}
\label{sec:limitations}

In this work, we have attempted to comprehensively address various common settings in which LLMs may be deployed for the task of Lay Summarisation. However, due to the inherent flexibility of prompt-based interaction, there will undoubtedly be variations that remain unexplored. Furthermore, it would have been desirable to assess the performance of an even wider range of models, particularly given the sheer number of LLM that are currently available and the rate at which they are released. However, in our model selection, we aim to ensure that we were assessing the performance of several widely-used models. Finally, in assessing the ability of models to generalise to non-biomedical articles, we could have selected any from a large number of viable domains. NLP was selected based on the high its rate of publication and high levels of interest it is currently attracting from lay audiences.

%% file: sections/A0.appendix0.tex
\begin{table*}[th]
    \centering
    \resizebox{0.9\textwidth}{!}{%
    \begin{tabular}{lm{0.90\textwidth}}
        \hline
        \textbf{Role} & \textbf{Prompt}  \\ \hline
          \multirow{6}{*}{\textbf{Author}} &  You are the author of the given research article tasked with answering the following questions about your work. When answering these questions please bear in mind that the audience for your answers is broad, and includes researchers from other fields (who will know relatively little about your own field of research) and interested members of the public. \\ 

         & \\
         & \textit{\{questions\}} \\
         & \\
         & \#\#\# ARTICLE \\
         & \textit{\{article\}} \\
         
         % \textit{{Questions}}\\
         \hline
         \textbf{Writer} & You are a freelance writer, tasked with summarising a biomedical research article for a lay audience. In addition to the article itself, the authors have answered a short questionnaire about their work. Using both the article text and the author-provided answers, summarize the article for a non-expert audience. Your summary should be between 300 and 400 words and contain minimal jargon, often using words and phrases that aren't present in the article. The first half of your summary should focus on explaining the background information that a lay audience will require, and the second half should explain the key experiments and results, finishing with a concluding sentence about the significance of the article. \\
         & \\
         & \#\#\# ARTICLE \\
         & \textit{\{article\}} \\        
         & \\
         & \#\#\# ANSWERS \\
         & \textit{\{question\_answers\}} 
         % \\ 
        %  \hline
        %  Editor & You are an expert editor, tasked with finalising drafted research article summaries that are aimed at a non-expert audience. The final edited summary should be between 300 and 400 words and contain minimal jargon, often using words and phrases that aren't present in the article. The first half of the summary should focus on explaining the background information that a lay audience will require, with the second half explaining the key experiments and results and finishing with a concluding sentence about the significance of the article. Finalise the following summary drafts: 
        % % \textit{{Input-output examples}}
        \\ \hline
    \end{tabular}
    }
    \caption{Prompts provided to for each LLM role in the proposed two-stage framework.}
    \label{tab:agent_response_prompts}
\end{table*}

\section{Prompts} \label{sec:appendix1}
Here we provide full details of the prompts used in each stage of our methodology. 

\paragraph{Base model prompt} As mentioned in \S\ref{sec:experiment1}, for our base approach, we utilise the prompt ``Generate a summary of the following article that is suitable for non-experts". This prompt was selected as it was found to obtain the best overall performance in a series of preliminary experiments, whereby we tested several variations of this instruction using different models.

\paragraph{Task decomposition prompts} Firstly, we provide the prompts used by in both stages of the proposed two-stage framework in Table \ref{tab:agent_response_prompts}. Note that, when providing these prompts to different LLMs, the structure and notation is sometimes altered slightly to conform with the recommended prompt template, but content remains the same.  

Within Table \ref{tab:agent_response_prompts}, variable text is represented as: \textit{\{variable\}}. To clarify the meaning of given variables, \textit{\{questions\}} refers to the questions and guidelines that eLife provides authors - we provide these verbatim in Table \ref{tab:questions}. \textit{\{article\}} refers to the input text from the article which is to be summarised (as covered in Appendix \ref{sec:appendix2}, we experiment with different techniques for input text selection). Finally, \textit{\{answers\}} refers to the answers generated by the LLM from the author question-answering stage. 

\paragraph{LLM evaluator prompt} Table \ref{tab:llm_eval_prompts} provides the prompt given the LLM judges in order to extract preference judgements. As mentioned in \S\ref{sec:experiment1}, to avoid positional bias, we randomly change the order in which the summaries generated by each approach are given, such that 50\% of each possible ordering is used for a given model.

\begin{table*}[]
    \centering
    \resizebox{1\textwidth}{!}{%
    \begin{tabular}{m{\textwidth}}
    \hline
    \textbf{Question and guidance} \\ \hline
1. What background information would someone who is completely unfamiliar with your field need to know to understand the findings in your paper? (Suggested word limit: 150 words)

- Include something that most readers will be able to relate to in the first sentence. Get gradually more specific in the following sentences.

- Don't try to explain the background to your entire field; instead consider which details a reader would need to know to understand the new findings, and then explain these facts as clearly and concisely as you can.

- Make sure to provide simple definitions or explanations for all technical terms and acronyms. \\ \hline

2. What exact research question did you set out to answer and why? (Suggested word limit: 75 words)

- Provide context by making it clear if this question was asking something completely new, or if you wanted to test or build upon previous findings.

- Make sure that you explain why it was important to find an answer this question (why should people care whether you can answer this question or not?). \\ \hline

3. What are the most important findings of your paper? (Suggested word limit: 100 words)
- Focus on findings highlighted in the title or abstract of your paper, and explain them clearly and completely.

- If possible, describe your methodology with a sentence or two.

- Always mention which species, type of organism or cells you have studied (for example, mutant mice, fruit flies, human kidney cells, or cancer cells). \\ \hline

4. Who might eventually benefit from the findings of your study, and what would need to be done before we could achieve these benefits? (Suggested word limit: 75 words)

- Think beyond your immediate field of research, and explain how your findings could lead to a benefit for wider society (patients, the environment, and so on).

- Avoid hype or exaggeration. For example, if your findings are about a fundamental process in living cells that could be relevant to understanding cancer, you should mention the link but be careful not to imply that the findings will imminently lead to new treatments. \\
\hline
    \end{tabular}
    }
    \caption{Questions and recommendations provided by eLife to accepted authors.}
    \label{tab:questions}
\end{table*}

\begin{table*}[th]
    \centering
    \resizebox{0.9\textwidth}{!}{%
    \begin{tabular}{m{0.90\textwidth}}
        \hline
        \textbf{Prompt}  \\ \hline
        You are a tasked with indicating your preference between two research article summaries that are intended for a non-expert audience. \\ \\
        
        Your preference should be based on which summary you believe would be more useful in informing a lay audience about the findings and significance of the article. \\ \\
        
        Respond with the number of the report you would recommend and a brief explanation of why you would recommend it.\\ \\

        \#\#\# SUMMARY 1 \\
        \textit{\{summary1\}} \\ \\
        
        \#\#\# SUMMARY 2 \\ 
        \textit{\{summary2\}} \\ \\

        \#\#\# PREFERENCE
        \\ \hline
    \end{tabular}
    }
    \caption{Preference judgement prompt.}
    \label{tab:llm_eval_prompts}
\end{table*}

%% file: sections/A1.appendix1.tex
\section{Additional Experimental Details} \label{sec:appendix2}
Here we provide additional details relating to experiments and results that are not given in the main text.

\paragraph{Dataset statistics}
The statistics relating to the length of the documents and summaries in the eLife dataset are provided in Table \ref{tab:datasets}.

\paragraph{Metric calculation}
ROUGE scores were calculated using the existing \href{https://pypi.org/project/rouge-score/}{\texttt{rouge-score}} package, with stemming and sentence tokenization applied
FKGL and DCRS were computed using the \href{https://github.com/shivam5992/textstat}{\texttt{textstat}} package.

\begin{table}[t]
    \centering
    \begin{tabular}{cccc} \hline
        \multirow{2}{*}{\textbf{\# Docs}} & \multicolumn{1}{c}{\textbf{Doc}} & \multicolumn{2}{c}{\textbf{Summary}} \\   \cline{2-4}
        & \# words   & \# words  & \# sents  \\  \hline
         4,828   & 7,806.1 & 347.6 & 15.7 \\
                \hline
    \end{tabular} 
    \caption{Average statistics of eLife dataset.} 
    \label{tab:datasets}
\end{table}

\begin{table*}[th]
    \centering
    \resizebox{0.85\textwidth}{!}{%
    \begin{tabular}{lccccccccccc} \hline
        \multirow{2}{*}{\textbf{Model}} & \multirow{2}{*}{\textbf{QA}} & \multirow{2}{*}{\textbf{\# Secs}} & \multicolumn{4}{c}{\textbf{Relevance}} && \multicolumn{2}{c}{\textbf{Readability}} && \multicolumn{1}{c}{\textbf{Factuality}} \\  \cline{4-7} \cline{9-10} \cline{12-12}
        &&& \textbf{R-1$\uparrow$} & \textbf{R-2$\uparrow$} & \textbf{R-L$\uparrow$} & \textbf{BeS$\uparrow$} && \textbf{FKGL$\downarrow$} & \textbf{DCRS$\downarrow$} && \textbf{BaS$\uparrow$} \\ 
                \hline
                Mixtral & \xmark & A & 45.59 & 11.02 & 42.85 & 84.17 && 14.08 & 9.69 && -3.06 \\
                Mixtral & \xmark & A+I & 43.79 & 9.95 & 41.11 & 83.45 && 15.13 & 10.10 && -3.09 \\
                Mixtral & \xmark & all & 40.65 & 9.09 & 37.99 & 83.33 && 15.01 & 10.43 && -3.46 \\
                Mixtral & \cmark & A & 45.49 & 10.41 & 42.90 & 83.95 && 13.92 & 9.57 && -3.12 \\
                Mixtral & \cmark & A+I &  44.93 & 10.19 & 42.31 & 83.76 && 14.35 & 9.67 && -3.05\\
                Mixtral & \cmark & all & 42.75 & 9.48 & 40.13 & 83.64 && 14.57 & 10.04 && -3.48 \\
                \hdashline
                LLAMA3 & \xmark & A & 45.59 & 11.22 & 43.03 & 84.72 && 10.37 & 8.43 && -3.34  \\
                LLAMA3 & \xmark & A+I & 45.32 & 11.89 & 42.51 & 84.92 && 11.59 & 9.01 && -3.04 \\
                LLAMA3 & \xmark & all & 16.98 & 2.63 & 15.68 & 67.02 && 58.08 & 10.45 && -5.43 \\
                LLAMA3 & \cmark & A & 44.90 & 10.68 & 41.93 & 84.93 && 11.99 & 9.34 && -3.27 \\
                LLAMA3 & \cmark & A+I & 46.38 & 11.47 & 43.43 & 85.03 && 12.47 & 9.37 && -3.18  \\
                LLAMA3 & \cmark & all & 16.23 & 2.48 & 14.69 & 66.72 && 62.96 & 10.15 && -5.23 \\
                \hline
    \end{tabular}}
    \caption{The effect of input selection on the performance of models on eLife test split. \textbf{R} = ROUGE F1, \textbf{BeS} = BERTScore F1, \textbf{FKGL} = Flech-Kincaid Grade Level, \textbf{DCRS} = Dale-Chall Readability Score,
    \textbf{BaS} = BARTScore.
    }
    \label{tab:input_selection}
\end{table*}

\paragraph{Input selection experiment}
We perform additional experiments whereby we seek to identify which input format provides the best performance.
Previous Lay Summarisation works have taken somewhat varied approaches when it comes to input selection. One common approach is to attempt to utilise the input article in full, or at least truncated at the maximum context length of the selected model \citep{goldsack-etal-2022-making, goldsack-etal-2023-enhancing}. However, past works have also demonstrated good performance using only the article abstract as input \citep{turbitt-etal-2023-mdc, Guo2020-ba}, minimising the contextual burden placed upon the model. 

Table \ref{tab:input_selection} offers a comparison of multiple methods of input selection for the Mixtral and LLAMA3 model using automatic metrics. Specifically, we experiment with abstract-only (A), abstract+introduction (A+I), and full-article (full). 
Interestingly, our results show that using only the abstract as input yields the best overall performance in the majority of cases, with the exception of LLAMA3 utilising our two-stage framework. Therefore, in our primary experiments, we utilise only the abstract as the article text for all models, in order to ensure a fair comparison.  

We also find that models that utilise the full article perform considerably worse than their equivalents with shorter inputs, indicating that content at the start of the article is significantly more relevant for lay summary production, and that zero-shot models may struggle to identify relevant content when presented with the full article. LLAMA was found to particularly struggle with the full article input, being unable to follow the instructions provided and produce a coherent output.

\begin{table*}[t]
    \centering
    \resizebox{0.85\textwidth}{!}{%
    \begin{tabular}{lccccccccc} \hline
        \multirow{2}{*}{\textbf{Model}} & \multicolumn{4}{c}{\textbf{Relevance}} && \multicolumn{2}{c}{\textbf{Readability}} && \multicolumn{1}{c}{\textbf{Factuality}} \\  \cline{2-5} \cline{7-8} \cline{10-10}
        & \textbf{R-1$\uparrow$} & \textbf{R-2$\uparrow$} & \textbf{R-L$\uparrow$} & \textbf{BeS$\uparrow$} && \textbf{FKGL$\downarrow$} & \textbf{DCRS$\downarrow$} && \textbf{BaS$\uparrow$} \\ 
                \hline
                Mixtral & 45.49 & 10.41 & 42.90 & 83.95 && 13.92 & 9.57 && -3.12 \\
                Mixtral$_{no\_guides}$ & 44.41 & 9.81 & 41.71 & 83.81 && 14.67 & 9.67 && -3.12 \\
                Mixtral$_{no\_roles}$ & 43.33 & 9.67 & 40.78 & 83.36 && 14.54 & 9.69 && -3.03 \\
                Mixtral$_{single\_prompt}$ & 44.29 & 10.13 & 41.49 & 84.18 && 14.24 & 9.87 && -3.15  \\
                \hdashline
                LLAMA3 & 45.59 & 11.22 & 43.03 & 84.72 && 10.37 & 8.43 && -3.34  \\
                LLAMA3$_{no\_guides}$ & 44.57 & 10.41 & 41.68 & 84.72 && 12.90 & 9.68 && -3.28 \\
                LLAMA3$_{no\_roles}$ & 44.94 & 11.01 & 42.12 & 84.87 && 12.63 & 9.58 && -3.20 \\
                LLAMA3$_{single\_prompt}$ & 45.81 & 10.41 & 43.25 & 84.08 && 13.03 & 9.28 && -3.74  \\
                % LLAMA2$_{AW}$ & 44.14 & 10.17 & 41.44 & 84.35 && 13.27 & 9.36 && -3.20 \\
                % LLAMA2$_{AW-guides}$ & 43.45 & 10.42 & 40.61 & 84.12 && 13.77 & 9.78 && -3.03 \\
                % LLAMA2$_{AW-roles}$ & 41.98 & 9.84 & 39.19 & 83.86 && 14.60 & 9.97 && -2.98 \\
                % LLAMA2$_{W}$ & 34.49 & 8.24 & 31.87 & 83.59 && 12.66 & 9.95 && -3.64 \\
                % \hline
                % \hline
                \hline

    \end{tabular}}
    \caption{Average performance of ablated versions of the two-stage models on eLife test split. \textbf{R} = ROUGE F1, \textbf{BeS} = BERTScore F1, \textbf{FKGL} = Flech-Kincaid Grade Level, \textbf{DCRS} = Dale-Chall Readability Score,
    \textbf{BaS} = BARTScore.
    }
    \label{tab:ablation_study}
\end{table*}

\paragraph{Ablation study experiment}
% - Editor has little contribution to overall performance
% - roles are quite important to the overall performance
% - guidelines are quite important to the overall performance

To determine the efficacy and contribution of each aspect of our two-stage framework, we perform an ablation study whereby we systematically remove various prompt components. Again, we utilise Mixtral and LLAMA 3 as the base models for this experiment, and measure model performance using automatic metrics. The results of this evaluation are presented in Table \ref{tab:ablation_study}; and we divide our discussion based on the element removed from the system:

\begin{itemize}[]
    \item[] \textit{Expert guidelines} - The removal of our expert-derived guidelines from the instruction prompts of the Writer is denoted by the LLM$_{no\_guides}$. Here, we observe a decrease in performance compared to the base model. 
    % However, this is significantly more severe in the case of LLAMA2$_{AWE-guides}$, providing further indication that this model struggles when left only with input-output examples. 
    
    % \item \textit{Editing stage} -
    \item[] \textit{Role descriptions} - We also experiment with the removal of role descriptions from the prompts of the Author and Writer stages which has shown to be significant in previous work \citep{wang2024unleashing, chan2023chateval}. In line with their findings, both of the LLM$_{no\_roles}$ model exhibits a surprisingly large drop in performance compared to the base models, attesting to the importance of role-playing in a setting where LLMs are used to represent different actors. 

    \item[] \textit{Multi-stage format} - Finally, the LLM$_{single\_prompt}$ model in Table \ref{tab:ablation_study} represents a final experiment whereby the Author stage is effectively removed, and the Writer is asked to answer the author questions before composing the lay summary within a single prompt. Again, we observe worse overall performance compared to the two-stage setting. 
\end{itemize}

\paragraph{Inter-annotator agreement}
Here we discuss the inter-annotator agreement by both human and LLM judges for the preference-based evaluation in Table \ref{tab:results_elife} and the in-depth evaluation in Figure \ref{fig:human_eval_results}.

Starting with the preference-based evaluation (Table \ref{tab:results_elife}) on biomedical articles, we measure both Cohen's $\kappa$ and the overall percentage of agreement obtained between pairs of annotators across all 100 instances (i..e, combining the 20 samples summaries for 5 the different models). For human evaluators, who's preferences are aggregated to obtain the PoH metric, we find that pairs of evaluators agree 55.17\% of the time and get an average $\kappa$ of 0.103 (across 4 lay evaluators). This corroborates the findings of \citet{goyal2022}, who identify an inherent variance the values of annotators when it comes to indentifying which summary they consider to be the ``best" or, in our case, ``most useful". Interestingly, we see a similar pattern for pairs of LLM judges, who we found to agree on 46.67\% of instances, and obtain an average  $\kappa$ of -0.067. 
Overall, these results further attest to the utility of having a panel of evaluators in a preference-based evaluation setting, due to the variance in the individual preferences of both humans and LLMs. Furthermore, this provide further evidence for ability of LLMs to mimic humans when evaluating in this setting, with our results showing that, although disagreement between individual evaluators is likely, the overall behavoir and preferences are largely similar to that of human judges.

For our question-based human evaluation Mixtral variants for NLP articles (Figure), we again measure the inter-annotator agreement using Cohen's $\kappa$ between expert evaluators. For this, we get an average pairwise score of 0.334 which, notably, increases to 0.428 if we combine the respective ``agree" and ``disagree" variables, attesting to the reliability of our evaluation.

\begin{table*}[th]
    \centering
    \resizebox{0.85\textwidth}{!}{%
    \begin{tabular}{lccccccccccc} \hline
        \multirow{2}{*}{\textbf{Model}} & \multicolumn{4}{c}{\textbf{Relevance}} && \multicolumn{2}{c}{\textbf{Readability}} && \multicolumn{1}{c}{\textbf{Factuality}} \\  \cline{2-5} \cline{7-8} \cline{10-10}
        & \textbf{R-1$\uparrow$} & \textbf{R-2$\uparrow$} & \textbf{R-L$\uparrow$} & \textbf{BeS$\uparrow$} && \textbf{FKGL$\downarrow$} & \textbf{DCRS$\downarrow$} && \textbf{BaS$\uparrow$} \\ 
                \hline
                BART & 46.57 & 11.65 & 43.70 & 84.94 && 10.95 & 9.36 && -2.39   \\ 
                Longformer & 47.23 & 13.20 & 44.44 & 85.11 && 11.88 & 9.09 && -2.56 \\  
                \hline
    \end{tabular}}
    \caption{Average performance of fine-tuned LMs on eLife test split. \textbf{R} = ROUGE F1, \textbf{BeS} = BERTScore F1, \textbf{FKGL} = Flesch-Kincaid Grade Level, \textbf{DCRS} = Dale-Chall Readability Score, \textbf{BaS} = BARTScore}
    \label{tab:results_elife_ft}
\end{table*}

\paragraph{Comparison to fine-tuned LMs} Table \ref{tab:results_elife_ft} presents the automatic metrics scores obtained by BART \citep{lewis-etal-2020-bart} and Longformer \citep{beltagy2020longformer} models that have been fine-tuned on the eLife training set, with these being the two LMs that have been most widely used in previous Lay Summarisation works. By comparing these scores to those obtained by zero-shot LLMs (e.g., in Table \ref{tab:results_elife}), we can see that the current generation of models is able to obtain almost comparable performance in most metrics, despite not having seen any training examples. Overall, this further illustrates the strong utility that current LLM have for this task, particularly when considered alongside the fact that (as we have shown in this work) zero-shot LMs are also able to generalise to previously unseen domains.

\paragraph{Additional results for NLP domain} Table \ref{tab:results_acl_other} presents the metric scores of the smaller LLMs (Phi and Mistral) for articles in the NLP domain.

\begin{table}[t]
    \centering
    \resizebox{\columnwidth}{!}{%
    \begin{tabular}{lccccccc} \hline
        \multirow{2}{*}{\textbf{Model}} &  \multirow{2}{*}{\textbf{QA}} & \multicolumn{2}{c}{\textbf{Readability}} && \multicolumn{1}{c}{\textbf{Factuality}} && \multicolumn{1}{c}{\textbf{H2H}} \\  \cline{3-4} \cline{6-6} \cline{8-8} 
        & & \textbf{FKGL$\downarrow$} & \textbf{DCRS$\downarrow$} && \textbf{BaS$\uparrow$} && \textbf{PoLL$\uparrow$} \\ 
                \hline
                 \multirow{2}{*}{Phi2} & \xmark & 14.05 & 8.21 && -3.25 && 0.75 \\
                & \cmark & 13.40 & 7.62 && -3.77 && 0.25 \\
                \hdashline
                \multirow{2}{*}{Mistral} & \xmark & 13.46 & 9.89 && -3.26 && 0.35 \\                
                & \cmark & 14.37 & 10.03 && -3.78 && 0.65 \\

                \hline
    \end{tabular}}
    \caption{Average performance of smaller models on ACL paper set. \textbf{FKGL} = Flech-Kincaid Grade Level, \textbf{DCRS} = Dale-Chall Readability Score,  
    \textbf{BaS} = BARTScore.
    }
    \label{tab:results_acl_other}
\end{table}

%% file: sections/A2.appendix2.tex
\section{Expanded related work} \label{sec:appendix3}

Automatic Lay Summarisation is a task that has started to attract increased attention in recent years, with the vast majority of attempts focusing exclusively on the biomedical domain. The one exception to this rule is the first attempt at the task, the LaySumm subtask of the CL-SciSumm 2020 shared task series \citep{Chandrasekaran2020-df} in which 8 teams participated, based around a corpus derived from Elsevier journal articles. 

Since then, \citet{Guo2020-ba} have experimented with applying then-state-of-the-art summarisation models to a novel dataset of biomedical systematic reviews paired with corresponding lay summaries. Similarly, \citet{goldsack-etal-2022-making} introduce, analyse, and benchmark two new datasets, PLOS and eLife (used in this work), which are derived from different biomedical journals of the same name. More recently, \citet{goldsack-etal-2023-enhancing} explored the incorporation of external knowledge graphs into Lay Summarisation models, demonstrating their potential to improve the readability of generated text.

Finally, the recent BioLaySumm shared task \citep{goldsack-etal-2023-biolaysumm} introduced some of the first attempts at Lay Summary generation using LLMs. \citet{turbitt-etal-2023-mdc} propose the winning approach, utilising GPT-3 (\texttt{text-davinci-003}) with the maximum number of in-context examples that can fit within the context window. Also notable is the submission of \citep{sim-etal-2023-csiro} that used ChatGPT for dataset augmentation through the generation of paraphrased references.  
We build on these studies, introducing a novel multi-stage methodology for Lay Summarisation and offering a comprehensive assessment of LLM performance across various dimensions to inform future research.